\documentclass[review]{elsarticle}
\usepackage{bbm}
\usepackage{mathrsfs}
\usepackage{times}
\usepackage{helvet}
\usepackage{courier}
\usepackage{graphicx}
\usepackage{amssymb}
\usepackage{multirow}
\usepackage{CJKutf8}
\usepackage{url}
\usepackage{latexsym}
\usepackage{amsmath}
\usepackage{lmodern}

\journal{arXiv}

\begin{document}
\begin{CJK*}{UTF8}{gbsn}

\begin{frontmatter}

\title{Neural Recovery Machine for Chinese Dropped Pronoun}

\author{Wei-Nan Zhang}
\ead{wnzhang@ir.hit.edu.cn}
\author{Ting Liu}
\author{Qingyu Yin}
\author{Yu Zhang}
\address{Research Center for Social Computing and Information Retrieval, School of Computer Science and Technology, Harbin Institute of Technology, Harbin, China, 150001}


\begin{abstract}
  Dropped pronouns (DPs) are ubiquitous in pro-drop languages like Chinese, Japanese etc.
  Previous work mainly focused on painstakingly exploring the empirical features for DPs recovery.
  In this paper, we propose a neural recovery machine (NRM) to model and recover DPs in Chinese, so that to avoid the non-trivial feature engineering process.
  The experimental results show that the proposed NRM significantly outperforms the state-of-the-art approaches on both two heterogeneous datasets.
  Further experiment results of Chinese zero pronoun (ZP) resolution show that the performance of ZP resolution can also be improved by recovering the ZPs to DPs.
\end{abstract}

\begin{keyword}
\texttt {neural network\sep dropped pronoun recovery\sep Chinese zero pronoun resolution}
\end{keyword}

\end{frontmatter}


\section{Introduction}

One of the key challenges in natural language understanding is to effectively model missing elements which are in some sense pragmatically inferable, such as dropped pronouns.
A segment of Chinese sentences observed in the real human to human dialogue text is shown in Table~\ref{tab1}.
The pronouns in square brackets are dropped in the Chinese sentences.
Humans can easily understand the meaning of the dialogue text due to the ``coherent model'' in mind.
However, it is non-trivial for the computer to understand the correct meaning of the incomplete sentences without recovering the dropped pronouns.
Therefore, it is one of the key steps to recover the dropped pronouns for elliptical sentence completion and natural language understanding for machines.
Recently, recovering the dropped pronouns has been verified to be effective for statistical machine translation~\cite{24}.

\begin{table}\small
\caption{A segment of Chinese dialogue text and its English translation. The words in square brackets are dropped pronouns which are omitted in the original text. CN, w2w and EN denote the Chinese sentence, word to word translation and English sentence corresponding to the CN, respectively.\label{tab1}}
\begin{center}{
\begin{tabular}{p{1.3cm}|l|l}
\hline
\multirow{3}{1in}{Human A:}
&CN& [你] 听见 了吗？\\
&w2w& [you] hear ? \\
&EN & Did you hear? \\
\hline
\multirow{3}{1in}{Human A:}
&CN & 他 说 [他] 要 买 个 Iphone 6s。 \\
&w2w & He say [he] will buy an Iphone 6s. \\
&EN & He says he will buy an Iphone 6s. \\
\hline
\multirow{3}{1in}{Human B:}
&CN & 嗯，[它] 很 贵 吧？\\
&w2w & uh, [it] very expensive ? \\
&EN & Is it very expensive? \\
\hline
\end{tabular}}
\end{center}
\end{table}

To address the dropped pronoun recovery problem, ~\cite{1} manually annotated dropped pronouns in Chinese newswire text.
~\cite{19} utilized a maximum entropy (ME) classifier to recover dropped pronoun from Chinese text messages.
However, the major drawbacks of the previous work are two-fold. First, manually annotating the dropped pronouns is a labor intensive work. Second, the applying of the ME classifier also dependents on empirically selecting features, which is also called the feature engineering.

In this study, we proposed a neural recovery machine (NRM) to recover the dropped pronouns on two heterogeneous datasets so that to avoid the non-trivial feature engineering process.
The contributions of this paper include the following:
\begin{itemize}
  \item To our knowledge, we are the first to propose a neural network framework, which avoids the empirical feature engineering process, to recover Chinese dropped pronouns.
  \item The proposed approach outperforms the state-of-the-art approach significantly on two heterogeneous datasets, which also demonstrate the domain adaption of the NRM.
  \item By integrating the dropped pronouns into the anaphoric zero pronouns, we verified that the proposed NRM can further improve the performance of zero pronoun resolution task.
\end{itemize}

\section{Preliminary}
\label{pre}
\subsection{Data Description}
\label{dd}
First, we utilized an \textbf{interview dialogue} from the ``bc'' section of the OntoNotes Release 4.0\footnote{http://catalog.ldc.upenn.edu/LDC2011T03, the files are: phoenix\_0000-phoenix\_0011, msnbc\_0000, cnn\_0000-cnn\_0004, cctv\_0000-cctv\_0007, p2.5\_cmn\_0001-p2.5\_cmn\_0061} which contains 2,501 sentences.
Second, we collected a \textbf{question answering dialogue}\footnote{A question answering dialogue is constructed by multiple turns of asking and answering between the asker and answerer.} from Baidu Zhidao\footnote{http://zhidao.baidu.com/} which contains 11,160 sentences.
We use both of the two heterogeneous datasets for the experiment of Chinese dropped pronoun recovery.
Table~\ref{tab2} shows the statistics of the datasets for annotation.
\begin{table}[htbp]
\caption{Statistics of the datasets for annotation.\label{tab2}}
\begin{center}{
\begin{tabular}{l|l|l}
\hline
& \textbf{\# of sentences} & \textbf{\# of words}  \\
\hline
OntoNotes 4.0 & 2,501 & 69,626 \\
\hline
Baidu Zhidao & 11,160 & 171,729 \\
\hline
\end{tabular}}
\end{center}
\end{table}

\subsection{Dropped Pronoun Annotation}

\subsubsection{Actual Pronouns}

In this study, we follow the annotation scheme in~\cite{1}.
The annotation framework which is proposed by~\cite{1} includes 14 types of pronouns.
Of these 14 pronouns, 10 of them are \textbf{actual pronouns} that are commonly used in Chinese speech and writing.
The details of these pronouns by following the description of~\cite{1} are listed below:

\begin{table}[htbp]\small
\caption{Actual pronouns and their descriptions.\label{tab3}}
\begin{center}{
\begin{tabular}{l|l}
\hline
\textbf{Actual Pronoun} & \textbf{Description} \\
\hline
我(I) & First person singular. \\
我们(we) & First person plural. \\
你(you) & Second person singular. \\
你们(you) & Second person plural. \\
他(he) & Third person masculine singular. \\
他们(they) & Third person masculine plural. \\
她(she) & Third person feminine singular. \\
她们(they) & Third person feminine plural. \\
它(it) & Third person inanimate singular. \\
它们(they) & Third person inanimate plural. \\
\hline
\end{tabular}}
\end{center}
\end{table}

Example (1) demonstrates a context for the inanimate third person singular pronoun.

\begin{small}
\begin{tabular}{lr|lllll}
(1)& CN & 那 & 杨洋 & 据 & 你 & 知道 \\
& w2w & Well & Yang yang & base on & your & know \\
& CN & \emph{[它]} & 需要 & 多 & 长 & 时间？\\
& w2w & \emph{[it]} & need & how & long & time? \\
\hline
& EN & \multicolumn{5}{p{7cm}}{Well , Yang yang , based on your understanding , how long will it \textbf{take}?}
\end{tabular}
\end{small}

In this example, the one that is ``take'' is dropped in the Chinese sentence (CN).
To contrast the CN to its English translation (EN), we can see that the dropped pronoun should be referred to the pronoun `` 它(it)''.

\subsubsection{Abstract Pronouns}

The rest 4 types of pronouns are called \emph{abstract pronouns} that exist in Chinese but do not correspond to any specific Chinese words. However, they are expressible in non pro-drop languages such as English.
We detail them as follows:

\noindent \textbf{existential}:
An existential subject usually appears in front of a small number of ``existence'' verbs, such as `` 有''(to have) and `` 存在''(to exist) etc.
Example (2) shows an existential subject preceding the verb ``有''(to have).

\begin{small}
\begin{tabular}{p{0.2cm}r|p{1.5cm}p{0.5cm}p{0.8cm}p{0.8cm}p{0.5cm}}
(2) & CN & \emph{[existential]} & 有 & 这么 & 事儿 & 嗯 \\
& w2w & {[existential]} & have & such & thing & um \\
\hline
& EN & \multicolumn{5}{p{6cm}}{\textbf{There was} such an incident, um}\vspace{6pt}
\end{tabular}
\end{small}

\noindent \textbf{unspecified}: An unspecified subject occurs when there is no one (or anyone) that should be interpreted.
This type of subject can sometimes be translated to ``one'' or ``someone''.
Example (3) shows an unspecified subject preceding the verb ``生''(to live).

\begin{small}
\begin{tabular}{p{0.2cm}r|p{2cm}p{0.6cm}p{0.6cm}p{1cm}}
(3) & CN & \emph{[unspecified]} & 生 & 在 & 国际 \\
& w2w & {[Someone]} & live & in & international \\
& CN & 都会 & 的 & \multicolumn{2}{l}{台北市} \\
& w2w & capital & DE & Taipei &  \\
\hline
& EN & \multicolumn{4}{p{6cm}}{For people \textbf{living} in cosmopolitan Taipei}\vspace{6pt}
\end{tabular}
\end{small}

\noindent \textbf{event}: An event subject is usually the word, phrase or clause that has occurred in context, but has no need to be repetitively referred.
Example (4) shows an event subject which is dropped in an interrogative sentence.

\begin{small}
\begin{tabular}{p{0.2cm}r|p{0.8cm}p{0.6cm}p{0.6cm}p{1cm}p{0.5cm}p{0.8cm}}
(4) & CN & 妈妈 & 说：& 那 & 奶奶 & 住 & 我们 \\
& w2w & Mom & said: & that & grandma & lives & our \\
& CN & 家，& \multicolumn{2}{l}{\emph{[event]}} & \multicolumn{2}{l}{好 吗？} & \\
& w2w & home, & \multicolumn{2}{l}{[event]} & \multicolumn{2}{l}{okay?} & \\
\hline
& EN & \multicolumn{6}{p{6cm}}{Mom said: ``How about grandma stays in our home , is \textbf{that} okay ?''} \vspace{6pt}
\end{tabular}
\end{small}

\noindent \textbf{pleonastic}: A pleonastic subject is usually has no actual semantic meaning and its existence is only to satisfy the syntactic need for a subject.
Example (5) shows a pleonastic subject preceding the verb ``下雪''(snowfall).

\begin{small}
\begin{tabular}{p{0.2cm}r|p{1.2cm}p{1.8cm}p{1.2cm}p{2cm}}
(5) & CN & 因为 & \emph{[pleonastic]} & 下雪 & 对 \\
& w2w & Because & [pleonastic] & snowfall & toward \\
& CN & 这个 & 北京 & 的 & 路面 \\
& w2w & this & Beijing & DE & road surface \\
& CN & 影响 & 会 & 比较 & 大   \\
& w2w & affect & will & relative & great \\
\hline
& EN & \multicolumn{4}{p{7cm}}{Since \textbf{snowfall} would affect road conditions in Beijing to a rather large extent.}\vspace{6pt}
\end{tabular}
\end{small}

\subsection{Annotation Statistics}

\begin{table}[htbp]\small
\caption{Statistics of the annotated dropped pronouns in the two datasets.\label{tab4}}
\begin{center}{
\begin{tabular}{l|cc}
\hline
 & OntoNotes 4.0  & Baidu Zhidao \\
 \hline
  我(I)          & 3.62 & \textbf{31.49} \\
  我们(we)       & 5.91 & 0.28 \\
  你(you)        & 5.41 & \textbf{31.31} \\
  你们(you)      & 0.21 & 0.13 \\
  他(he)         & 6.92 & 1.38 \\s
  他们(they)     & \textbf{11.96} & 0.3 \\
  她(she)        & 2.80 & 0.63 \\
  她们(they)     & 0.18 & 0.00 \\
  它(it)         & \textbf{19.56} & \textbf{33.08} \\
  它们(they)     & 1.37 & 1.96 \\
  existential    & 6.12 & 0.00 \\
  unspecified    & \textbf{15.39} & 0.00 \\
  event          & 6.86 & 0.00 \\
  pleonastic     & 9.69 & 0.00 \\
\hline
\end{tabular}}
\end{center}
\end{table}

For data annotation, we follow the annotation scheme proposed by~\cite{1}.
To annotate the OntoNotes 4.0 sentences, two native Chinese speakers that do not participate in our experiment design are invited to annotate the DPs.
To facilitate the annotation and avoid the ambiguity, we asked the two annotators to take the English translations of the Chinese corpus, which are also offered in the OntoNotes release 4.0 data, as reference.
This is because English is a non pro-drop language, while the Chinese DPs usually exist in their corresponding English translations.
To annotate the Baidu Zhidao sentences, three native Chinese speakers that do not participate in our experiment design are invited to annotate the DPs.
The final labels are obtained by voting.
The annotating agreement measured by Cohen's \texttt{Kappa}~\cite{10} equals to 0.71, which denotes a ``good'' agreement.
The distribution of these types of dropped pronouns in our data set is shown in Table~\ref{tab4}.

\section{Our Approach}\label{app}

\begin{figure*}[htbp]
\centering
{\includegraphics[width=.5\linewidth]{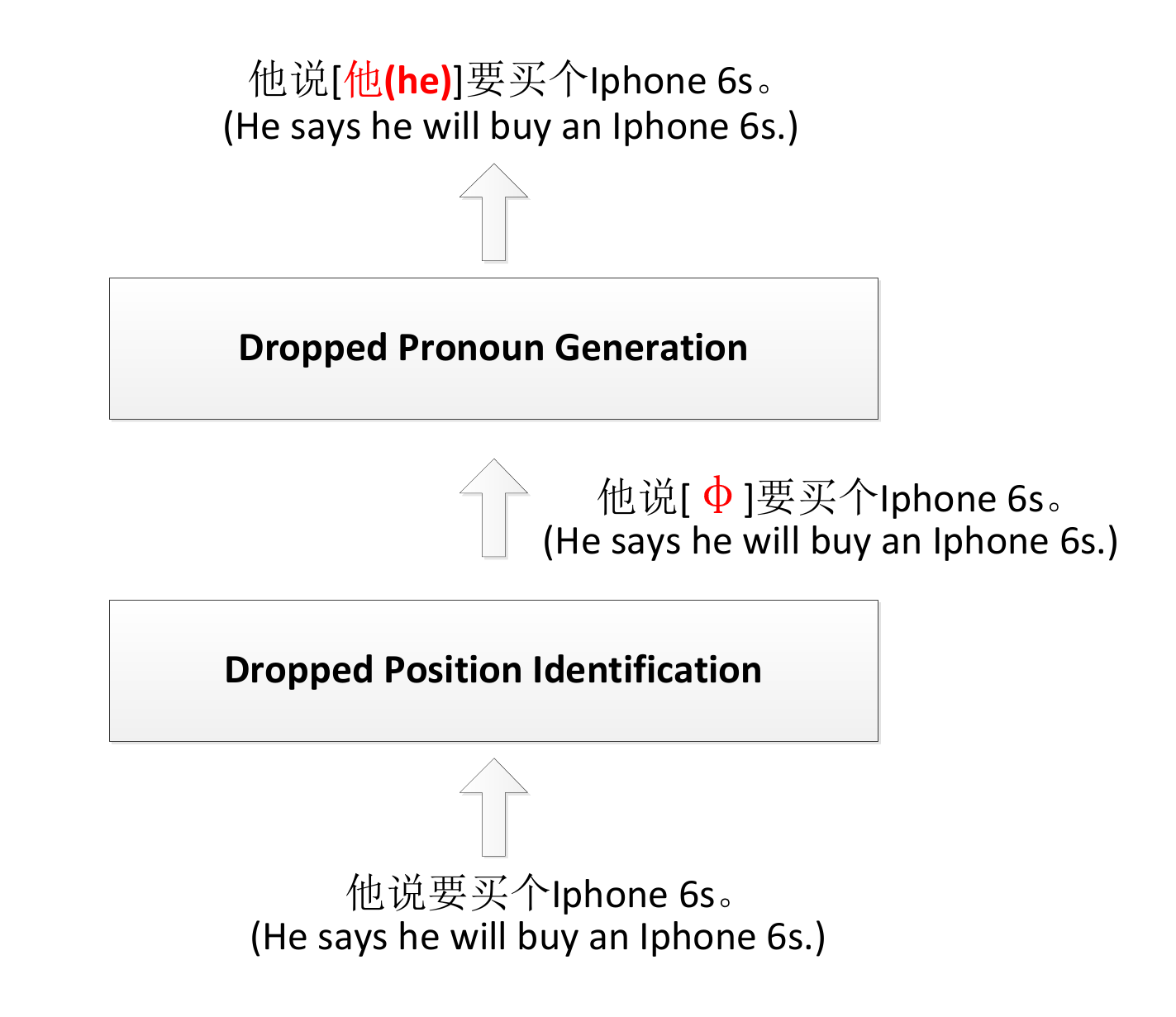}
\caption{The general framework of the proposed NRM for dropped pronoun recovery with an example.} \label{fig1}}
\end{figure*}

In this section, we detail the proposed NRM for dropped pronoun recovery.
Figure~\ref{fig1} shows the general framework of the NRM for dropped pronoun recovery with an example.

Here, the dropped position identification process focuses on choosing the position where a dropped pronoun should exist.
The dropped pronoun generation process then generates a explicit pronoun on the dropped position.

\subsection{Dropped Position Identification}\label{dpi}

\begin{figure*}[htbp]
\centering
{\includegraphics[width=1.0\linewidth]{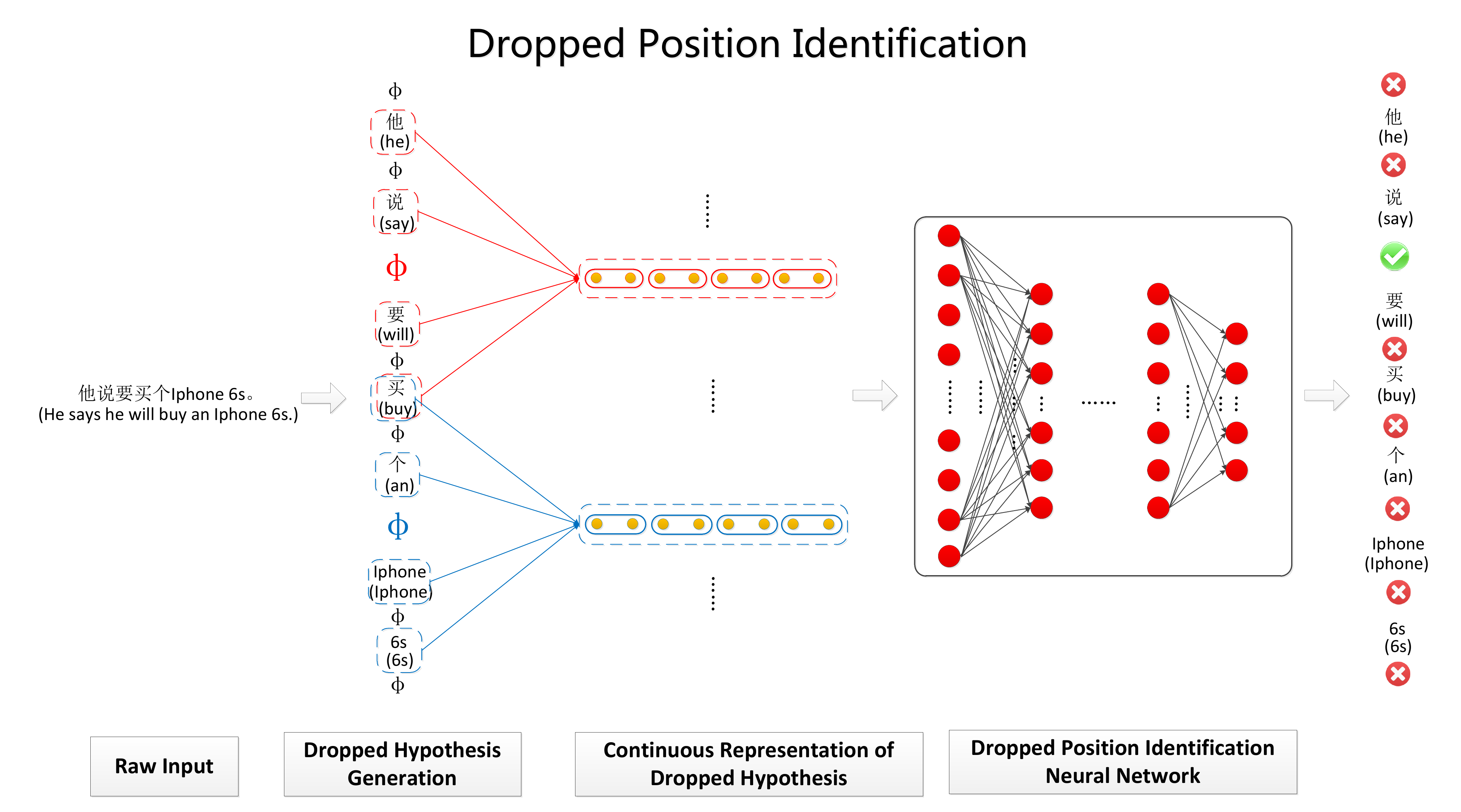}
\caption{The framework of the dropped position identification.} \label{fig2}}
\end{figure*}

Figure~\ref{fig2} shows the framework of the dropped position identification.
Here, given a raw input, we first carry out the preprocessing which is described in Section~\ref{preprocessing}.
To generate the dropped hypothesis ($\phi$), we assume that there may be a dropped position between each of the adjacent words as well as the beginning and the end of an input sentence as shown in Figure~\ref{fig2}.
Each of a possible dropped position is a dropped hypothesis.
To represent the dropped hypothesis, we utilize the context embedding which is constructed by concatenating the word embedding in a specific window.
We then utilize a neural network to identify the dropped position in a sentence.
Here, we use a multi-layer perceptron to realize the dropped position identification neural network.
Note that the dropped position identification neural network can be also realized by the convolutional neural network (CNN), the recurrent neural network (RNN), etc.
We will leave the exploration of these neural networks to the future work.

\subsection{Dropped Pronoun Generation}\label{dpg}

\begin{figure*}[!h]
\centering
{\includegraphics[width=.55\linewidth]{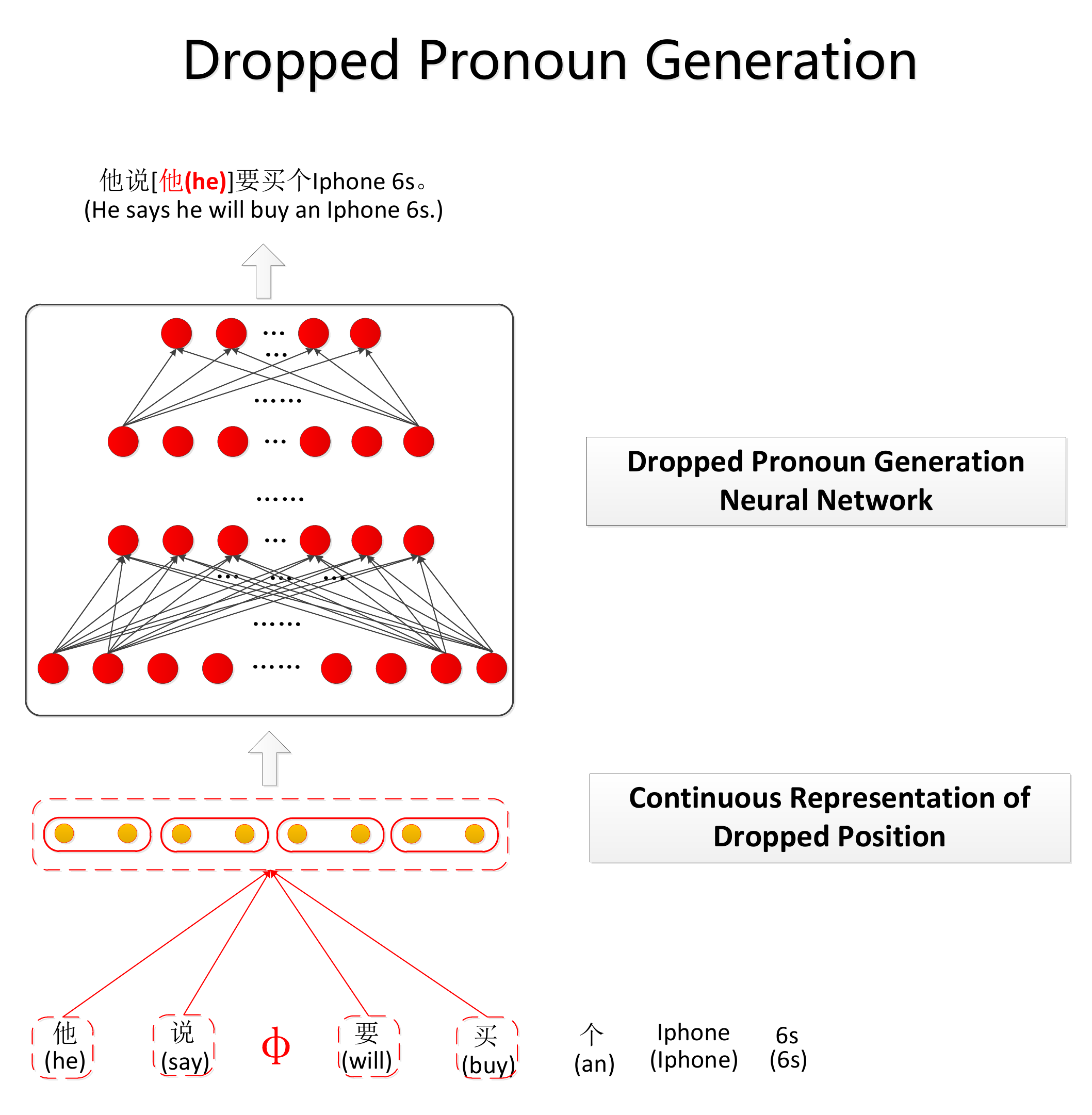}
\caption{The framework of the dropped pronoun generation.}\label{fig3}}
\end{figure*}

After the dropped position is fixed, the next process is to generate the explicit pronoun in a sentence.
Figure~\ref{fig3} shows the framework of the dropped pronoun generation.
As shown in Figure~\ref{fig3}, we represent the dropped position in a sentence by using its context embedding which is constructed by concatenating the word embedding in a specific window.
We then utilize a neural network to generate the explicit dropped pronoun for the dropped position.
The dropped pronoun generate neural network is also realized by using a multi-layer perceptron.

\subsection{Multi-Layer Perceptron Model}\label{mlp}

~\cite{21} proposed a unified framework for natural language processing (NLP), such as part-of-speech tagging, chunking, named entity recognition, semantic role labeling, language modeling, semantically related words, etc.
In this paper, we focus on another NLP task, named dropped pronoun recovery in the pro-drop language such as Chinese, Japanese, etc.
Inspired by~\cite{21}, to the specification of the dropped pronoun recovery task, we proposed a multi-layer perceptron model to realize the dropped position identification neural network (Section~\ref{dpi}) and the dropped pronoun generation neural network (Section~\ref{dpg}).
We use $x$, $W_i$ and $b_i$ to represent the continuous representation of the input, the weight matrix and bias of the $i$-th layer respectively.
Here, the continuous representation of the input is obtained by concatenating the word embedding, which is pre-trained as described in Section~\ref{preprocessing}, in a specific window in a sentence.
We use the ReLU as the activation function, which is described in Equation (\ref{eq1}).

\begin{equation}\label{eq1}
f_i = ReLU(W_{i}x+b_i)
\end{equation}
Here, we take $rect()$ as the $ReLU()$ and $z=W_{i}x+b_i$.
The activation function can be represented as Equation (\ref{eq2}).
\begin{equation}\label{eq2}
rect(z) = \max(z,0)
\end{equation}

As shown in Figure~\ref{fig1}, the dropped pronoun recovery includes 2 steps.
The first step is to identify the dropped position.
Hence, for each of the dropped hypothesis, the dropped position identification is transformed to a binary classification task.
The second step is to generate the explicit pronouns for the fixed dropped position.
It therefore can be seen as a multi-classification task.
According to the statistics of the annotated data which is shown in Table~\ref{tab4}, the dropped pronoun generation is transformed to a 14-category and a 10-category classification task on the OntoNote 4.0 and Baidu Zhidao datasets, respectively.
We thus utilize a $softmax$ function for the output layer.
The $softmax$ function is represented in Equation (\ref{eq3}).
\begin{align}\label{eq3}
\hat y &= softmax(W_{o}h+b_{o}) \\
\nonumber       &= \frac{\exp(\hat h^{T}w_i+b_i)}{\sum_{i^{\prime}}\exp(\hat h^{T}w_{i^{\prime}}+b_{i^{\prime}})}
\end{align}

The cross entropy is set to be the cost function as follows:
\begin{equation}\label{eq4}
C(y, \hat y) = -\sum_{i}y_{i}\log(\hat y_i)
\end{equation}
The stochastic gradient descent (SGD) is used to update the variables.

To avoid overfitting in training step, we employ the dropout regularization scheme after each activation function.

\section{Experimental Results}\label{exp}

\subsection{Data Preprocessing}\label{preprocessing}

To obtain the input of the NRM, there are two preprocessing steps.
The first is Chinese word segmentation that segments the input raw sentences into Chinese word sequences.
In this paper, we use the LTP cloud~\footnote{http://www.ltp-cloud.com/} service, which is a state-of-the-art Chinese word segmenter, to obtain the segmentation result.
Note that the OntoNotes 4.0 data has the manually word segmentation result so that the LTP cloud is only applied on the Baidu Zhidao data.
The second one is to transform the raw words into continuous word embedding.
Here, we utilize the Word2Vec toolkit\footnote{https://code.google.com/p/word2vec/} to obtain the Chinese word embedding.
The learning theory of the Word2Vec is described in~\cite{20}.
We train the Word2Vec with the SougouCS corpus (2008 version)\footnote{http://www.sogou.com/labs/dl/cs.html}, which is a Chinese newswire corpus of which size is 1.65GB.
LTP cloud is also used for the word segmentation of the corpus.

\subsection{Parameter Tuning}

For the experiments, we separate the data in Table~\ref{tab2} to training data, development data and test data in the proportion of $3:1:1$ in both of the OntoNotes 4.0 and Baidu Zhidao datasets.
We use the development data to tune the parameters of the proposed NRM.

\begin{figure}[htbp]
\centering
{\includegraphics[width=1.0\linewidth]{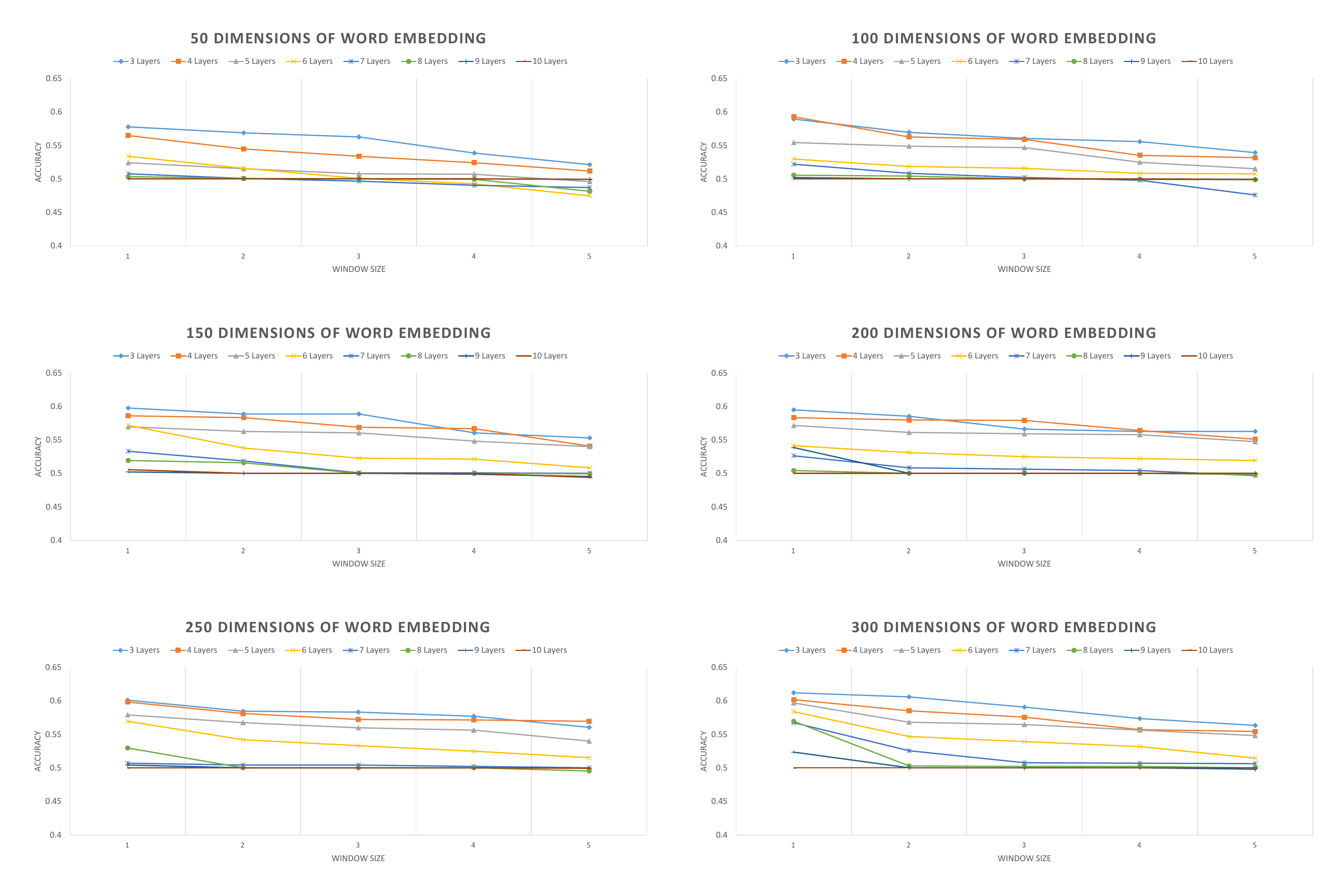}
\caption{The accuracy of the dropped position identification over the dimension of word embedding, the window size of the dropped hypothesis and the number of layers of the multi-layer perceptron on the OntoNotes 4.0 dataset.} \label{fig4}}
\end{figure}

\begin{figure}[htbp]
\centering
{\includegraphics[width=1.0\linewidth]{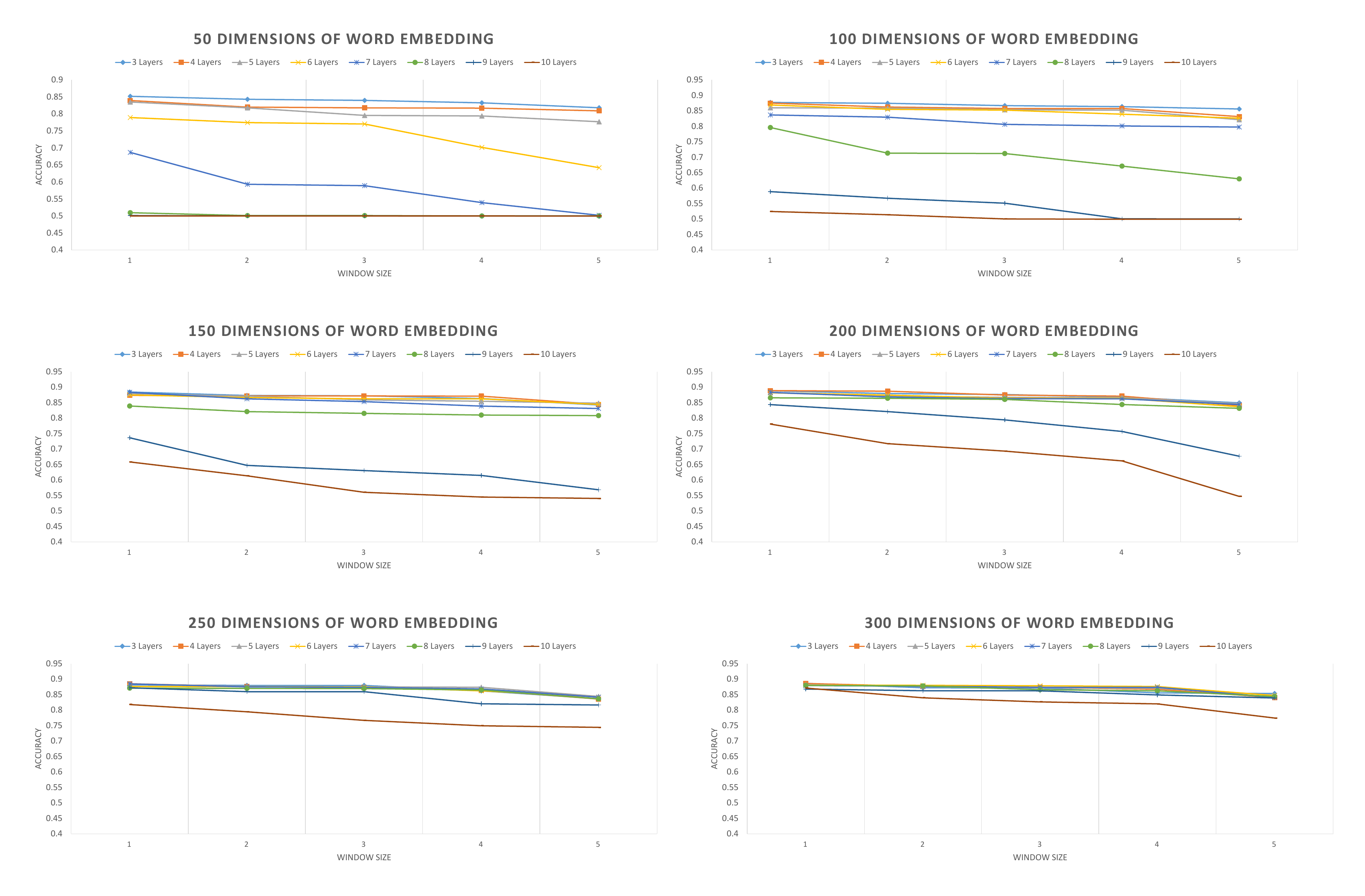}
\caption{The accuracy of the dropped position identification over the dimension of word embedding, the window size of the dropped hypothesis and the number of layers of the multi-layer perceptron on the Baidu Zhidao dataset.} \label{fig5}}
\end{figure}

\begin{figure}[htbp]
\centering
{\includegraphics[width=1.0\linewidth]{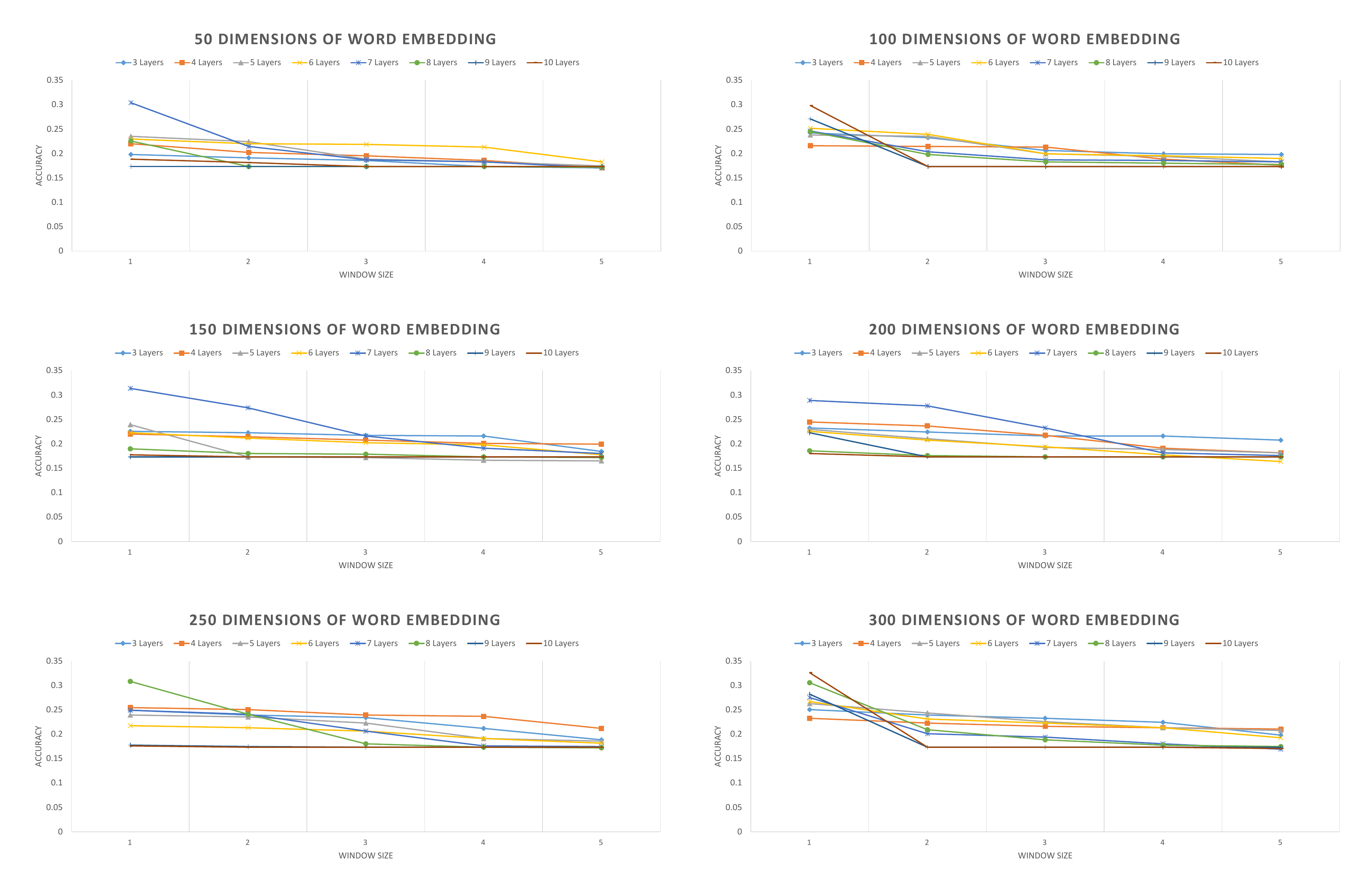}
\caption{The accuracy of the dropped pronoun generation over the dimension of word embedding, the window size of the dropped hypothesis and the number of layers of the multi-layer perceptron on the OntoNotes 4.0 dataset.} \label{fig6}}
\end{figure}

\begin{figure}[htbp]
\centering
{\includegraphics[width=1.0\linewidth]{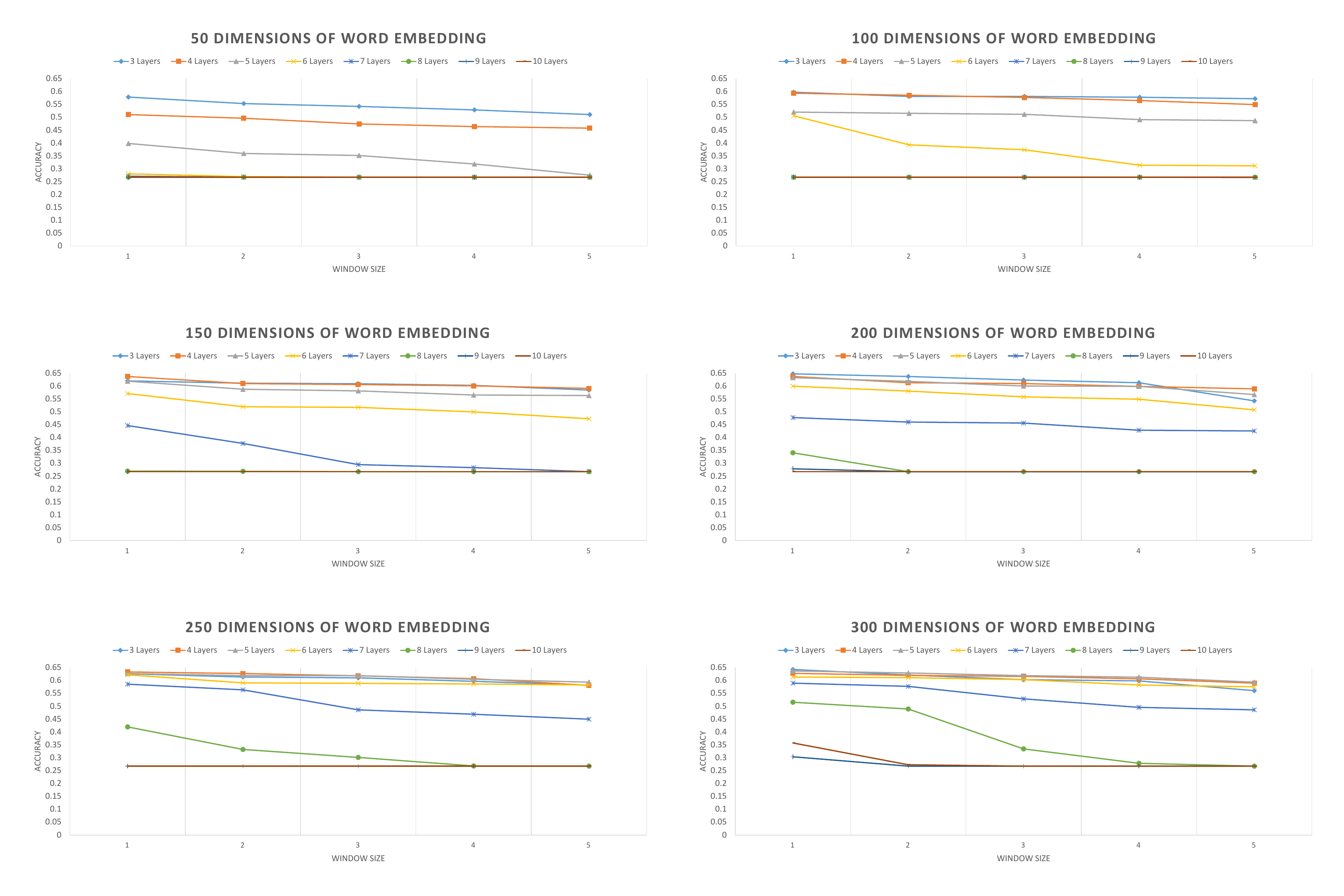}
\caption{The accuracy of the dropped pronoun generation over the dimension of word embedding, the window size of the dropped hypothesis and the number of layers of the multi-layer perceptron on the Baidu Zhidao dataset.} \label{fig7}}
\end{figure}

First, we consider the accuracy variation over three parameters.
They are \textbf{1)} the dimension of word embedding, \textbf{2)} the window size of the dropped hypothesis and \textbf{3)} the number of layers of the multi-layer perceptron for the dropped position identification and dropped pronoun generation, respectively.
Figure~\ref{fig4} and \ref{fig5} shows the accuracy of \textbf{dropped position identification} over the dimension of word embedding, the window size of the dropped hypothesis and the number of layers of the multi-layer perceptron on the OntoNotes 4.0 and Baidu Zhidao datasets, respectively. Figure~\ref{fig6} and \ref{fig7} show the accuracy of \textbf{dropped pronoun generation} over the dimension of word embedding, the window size of the dropped hypothesis and the number of layers of the multi-layer perceptron on the OntoNotes 4.0 and Baidu Zhidao datasets, respectively.

\begin{table}[htbp]
\caption{Parameter setting. $\mathbb{D}$ represents the dimension of word embedding. $\mathbb{W}$ represents the window size of the dropped hypothesis. $\mathbb{L}$ represents the number of layers of the multi-layer perceptron. $\mathbb{D}\mathbbm{o}$ represents the dropout parameter. $\mathbb{E}$ represents the epoch times. DPI and DPG represent the dropped position identification and dropped pronoun generation respectively. \label{tab_psetting}}
\begin{center}{
\begin{tabular}{l|cccc}
\hline
\multirow{2}{*}{} & \multicolumn{2}{c}{\textbf{DPI}} & \multicolumn{2}{c}{\textbf{DPG}} \\
\cline{2-5}
& \textbf{OntoNotes 4.0} & \textbf{Baidu Zhidao} & \textbf{OntoNotes 4.0} & \textbf{Baidu Zhidao} \\
\hline
$\mathbb{D}$ & $300$ & $200$ & $300$ & $200$ \\
$\mathbb{W}$ & $1$ & $2$ & $1$ & $2$ \\
$\mathbb{L}$ & $3$ & $4$ & $10$ & $3$ \\
$\mathbb{D}\mathbbm{o}$ & $0.5$ & $0.5$ & $0.8$ & $0.3$ \\
$\mathbb{E}$ & $25$ & $40$ & $10$ & $5$ \\
\hline
\end{tabular}}
\end{center}
\end{table}

\begin{figure}[htbp]
\centering
{\includegraphics[width=0.75\linewidth]{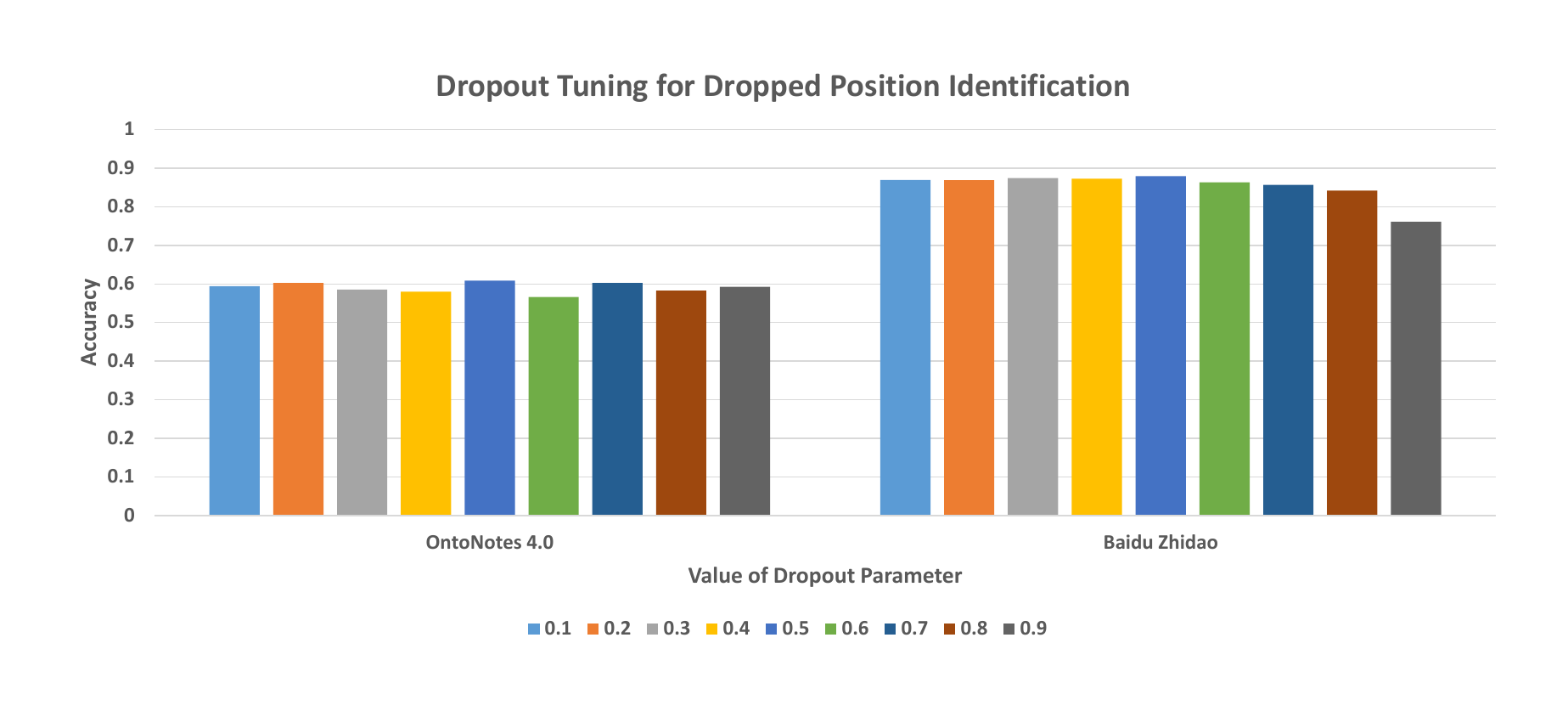}
\caption{The accuracy of the dropped position identification over the value of dropout parameter on the two datasets.} \label{fig8}}
\end{figure}

\begin{figure}[htbp]
\centering
{\includegraphics[width=0.75\linewidth]{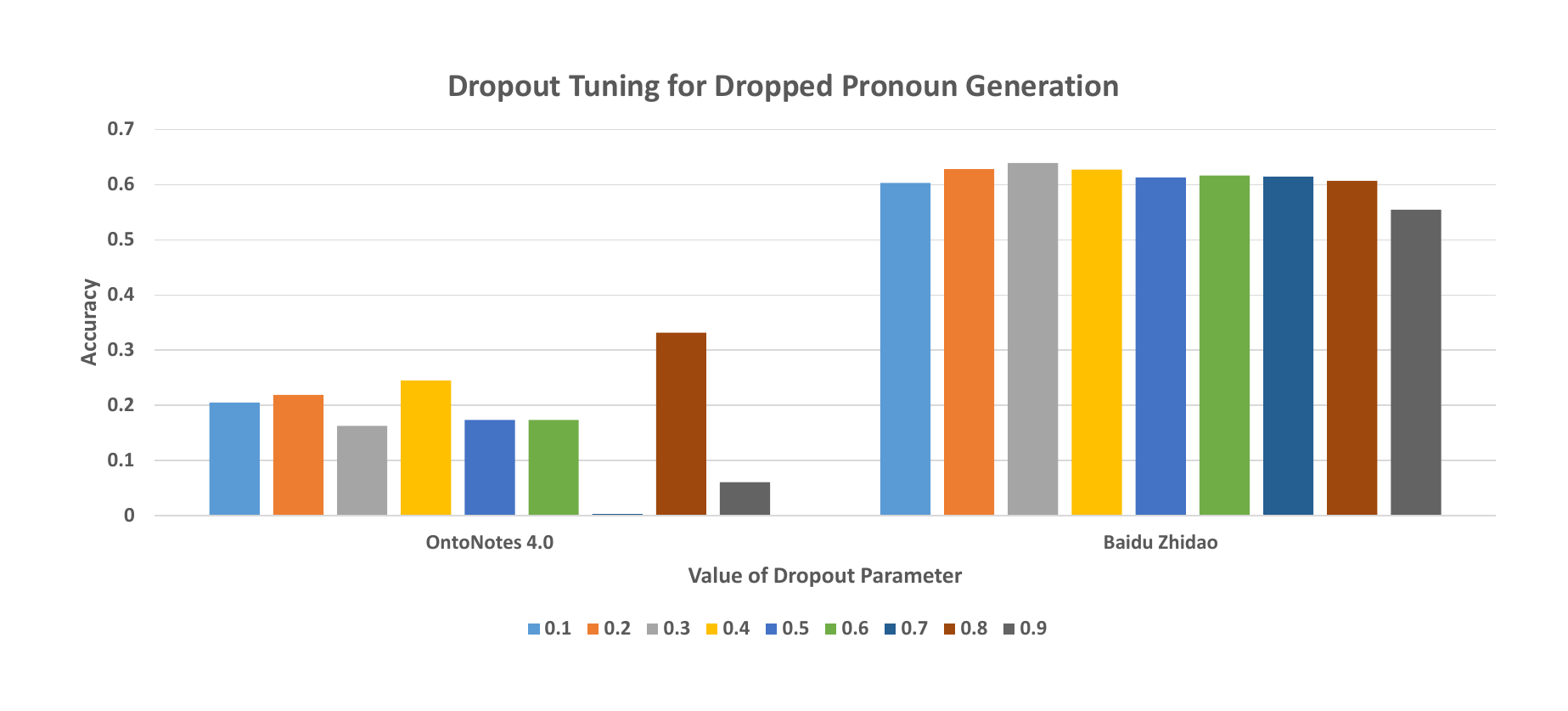}
\caption{The accuracy of the dropped pronoun generation over the value of dropout parameter on the two datasets.} \label{fig9}}
\end{figure}

Second, Figure~\ref{fig8} and \ref{fig9} show the accuracy variation of dropped position identification and dropped pronoun generation over \textbf{4)} the value of dropout parameter on the two datasets, respectively.

\begin{figure}[htbp]
\centering
{\includegraphics[width=0.75\linewidth]{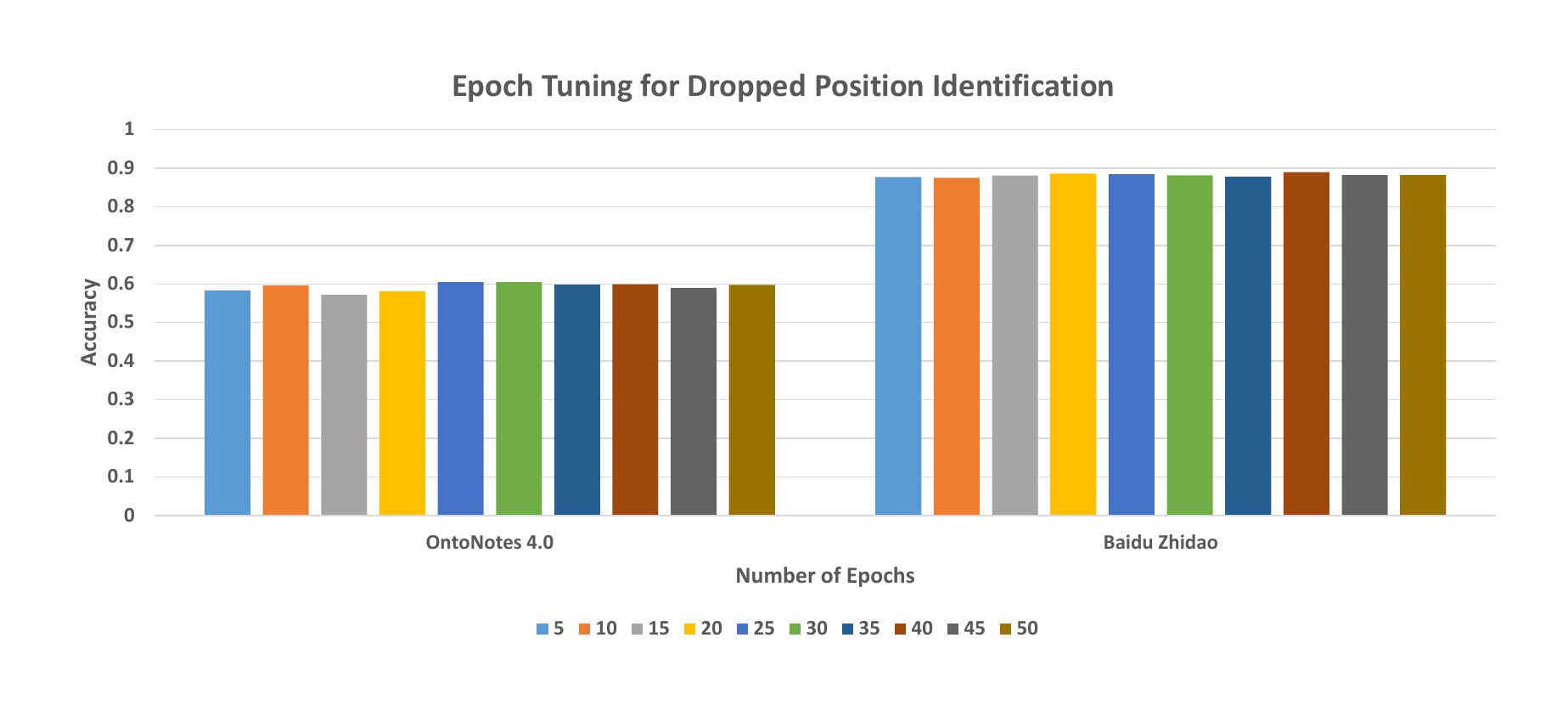}
\caption{The accuracy of the dropped position identification over the number of epochs on the two datasets.} \label{fig10}}
\end{figure}

\begin{figure}[htbp]
\centering
{\includegraphics[width=0.75\linewidth]{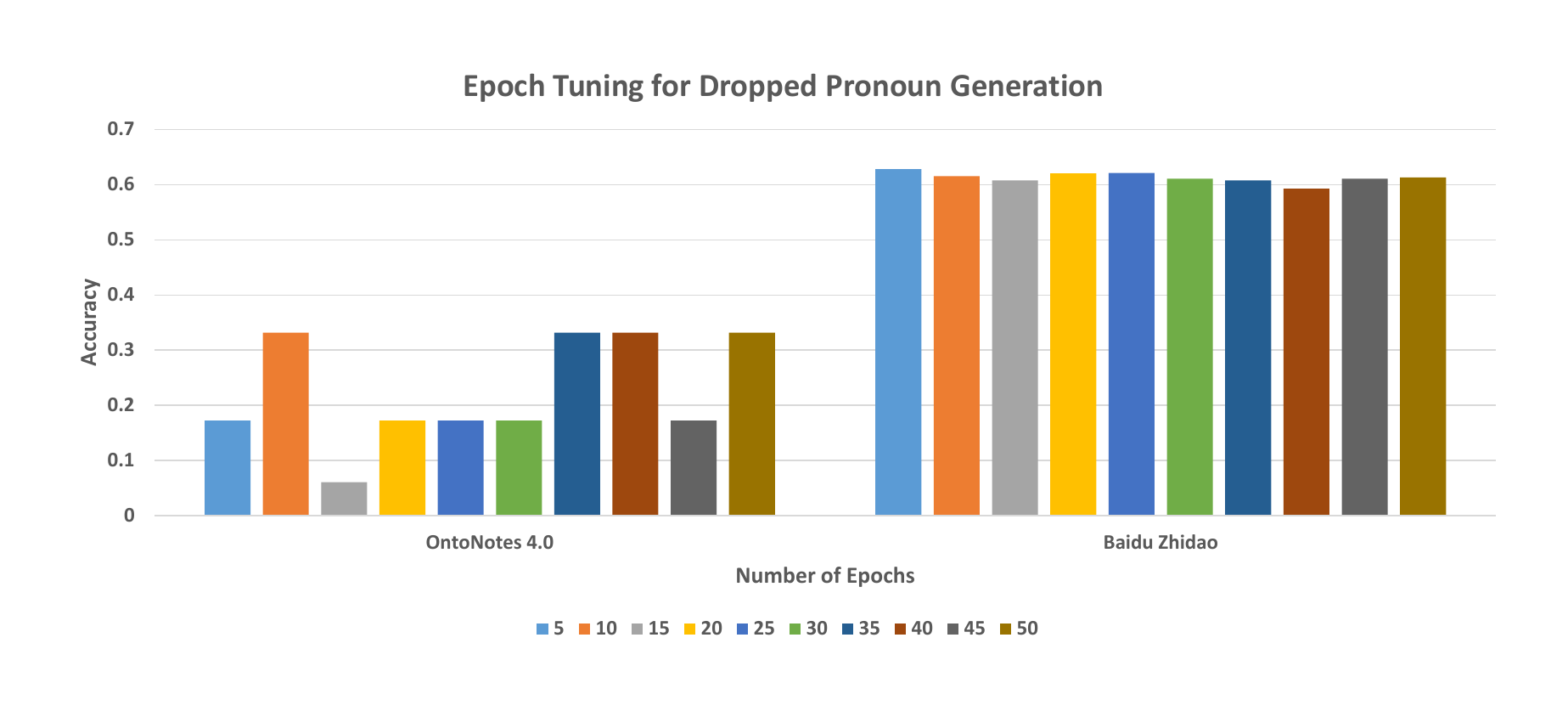}
\caption{The accuracy of the dropped pronoun generation over the number of epochs on the two datasets.} \label{fig11}}
\end{figure}

Third, Figure~\ref{fig10} and \ref{fig11} show the accuracy variation of dropped position identification and dropped pronoun generation over \textbf{5)} the number of epochs on the two datasets respectively.

After the parameter tuning, we get the parameter setting, which is shown in Table~\ref{tab_psetting}, for the experiments.

\subsection{Dropped Pronoun Recovery Results}

In our experiments, we use accuracy to evaluate the proposed approach.
Four comparison systems are selected as baselines.
They are SVM~\cite{25} with linear kernel and sparse vector input ($SVM_{SL}$), SVM with linear kernel and dense vector input ($SVM_{DL}$), SVM with sigmoid kernel and sparse vector input ($SVM_{SS}$) and SVM with sigmoid kernel and dense vector input ($SVM_{DS}$).
Meanwhile, we also re-implement a state-of-the-art (SOTA) approach to recovering dropped pronoun (Yang et al. 2015~\cite{19}) from Chinese text message for empirical comparison.
Here, the sparse vector is constructed by transforming the context of a dropped hypothesis into a vocabulary size vector.
When the words occur in the context, the elements in the corresponding positions of the vocabulary are set to 1 and the elements in other positions are set to 0.
The dense vector is produced by concatenating the embedding of the words in the context of a dropped hypothesis.

We compare the performance of the proposed NRM and the baselines in both of the OntoNotes 4.0 and Baidu Zhidao datasets.
The experimental results are shown in Tabel~\ref{tab5}.

\begin{table}[htbp]
\caption{Experimental results on dropped pronoun recovery. $\ast$ indicate that our approach is statistical significant over all the baselines (within $0.95$ confidence interval using the $t$-test).\label{tab5}}
\begin{center}{
\begin{tabular}{l|cccc}
\hline
\multirow{3}{*}{\textbf{Models}} & \multicolumn{4}{c}{\textbf{Accuracy}}  \\
\cline{2-5}
& \multicolumn{2}{c}{\textbf{DPI}} & \multicolumn{2}{c}{\textbf{DPG}} \\
\cline{2-5}
& \textbf{OntoNotes 4.0} & \textbf{Baidu Zhidao} & \textbf{OntoNotes 4.0} & \textbf{Baidu Zhidao} \\
\hline
$SVM_{SL}$ & $0.59$ & $0.755$ & $0.2$ & $0.456$ \\
$SVM_{SS}$ & $0.5$ & $0.5$ & $0.073$ & $0.315$  \\
$SVM_{DL}$ & $0.562$ & $0.739$ & $\textbf{0.21}$ & $0.47$ \\
$SVM_{DS}$ & $0.5$ & $0.5$ & $0.073$ & $0.315$ \\
SOTA & $0.524$ & $0.58$ & $0.191$ & $0.246$ \\
NRM & $\textbf{0.592}^{\ast}$ & $\textbf{0.879}^{\ast}$ & $0.18$ & $\textbf{0.584}^{\ast}$ \\
\hline
\end{tabular}}
\end{center}
\end{table}

We can see from Table~\ref{tab5} that our approach significantly outperforms the baselines in statistics on both of the two datasets for DPI task and the Baidu Zhidao dataset for DPG task.
To compare the results of the $SVM_{SL}$ and $SVM_{DL}$, we can see that the dense representation of the dropped hypothesis has better impact on the fine grained task, namely DPG rather than the DPI task.
To compare the results of $SVM_{SL}$ and $SVM_{SS}$ as well as the $SVM_{DL}$ and $SVM_{DS}$, we can see that the results of using linear kernel function are better than the use of sigmoid function\footnote{Here, we use the default setting of the parameters in LibSVM toolkit. The LibSVM toolkit is available at http://www.csie.ntu.edu.tw/~cjlin/libsvm }.
To see the results on the two datasets, we can see that the results on Baidu Zhidao dataset are better than those on OntoNotes 4.0 dataset.
It may be caused by the different scales of the two datasets and the different category numbers.
We also note that the SOTA, which is a feature engineering approach, performs good when the data scale is small, such as the experiment results of DPG on OntoNotes 4.0 dataset.
It may illustrates that the feature engineering approach can fit the data better when the data scale is small.
However, the feature engineering is a highly empirical process and also a complicated task.

\subsection{Error Analysis on Dropped Pronoun Recovery}

Further, one challenge on DPs recovery is the circumstance that the referents for DPs are clear, but the pronouns which represent them are not.
This commonly occurs with nouns that represent organizations.
An organization can be a referent that is referred by a singular or a plural.
For example, ``公司(company)'' can be the company itself as one unit or the people in the company, in which case the corresponding pronouns would be ``它(it)'' and ``他们(they)'', respectively.
Hence, this can be viewed as an ambiguity which is intricate to model.
The following two sentences are examples to show the above cases.


\begin{small}
\begin{tabular}{lr|llllllllll}
(7) & CN & 礼来 & 公司 & 在 & \multicolumn{2}{l}{新泽西州} & 吗 ？ & & & & \\
& w2w & Eli & company & at & \multicolumn{2}{l}{New Jersey} & MA ? & & & & \\
\hline
& EN & \multicolumn{10}{p{6cm}}{Is Eli Lilly in New Jersey?} \\
& \multicolumn{10}{p{6cm}}{} \\
& CN & 我 & 不 & 知道 & \emph{[他们]} & 公司 & 地址 & 在 & 哪里 。 &  & \\
& w2w & I & don't & know & \emph{[they]} & company & address & at & where . & & \\
\hline
& EN & \multicolumn{10}{p{6cm}}{I don't know where \textbf{they} are based.}
\end{tabular}
\end{small}

In example (7), the ``Eli Lilly'' is interpreted by the Chinese DP ``他们(they)''.

Besides that, the confusion caused by the gender of DPs is another challenge.
For instance, when the antecedents are absent on context, it is non-trivial to determine the explicit pronouns for recovery.
Example (8) shows the above case.

In example (8), the antecedent of the dropped pronoun ``她(She)'' is absent on context.
Other personal pronouns can also be placed in the position of the DP and are sensible both on syntactic and semantic.
This problem can be alleviated by the anaphora resolution technique, which will be exploited by us in future work.

\begin{small}
\begin{tabular}{lr|llllllll}
(8) & CN & \emph{[她]} & 看起来 & \multicolumn{2}{l}{少年老成 ，} & \multicolumn{2}{l}{是 吧 ？} & & \\
& w2w & \emph{[She]} & looks like & \multicolumn{2}{l}{young but mature ,} & \multicolumn{2}{l}{is it ?} & & \\
\hline
& EN & \multicolumn{8}{p{6cm}}{\textbf{She} seems young and competent, right?} \\
& \multicolumn{8}{p{6cm}}{} \\
& CN & \emph{[她]} & 真的 & 是 & 这样 。& & & & \\
& w2w & \emph{[She]} & really & is & this . & & & & \\
\hline
& EN & \multicolumn{8}{p{6cm}}{\textbf{She} really does.}\vspace{6pt}
\end{tabular}
\end{small}

\subsection{Zero Pronoun Resolution Results}

As the zero pronouns have no indicative information, such as the overt pronouns, we are thus motivated to recovering the dropped pronouns for the zero pronoun resolution task.
For integration, we first run our NRM on the OntoNotes 5.0 data, which is a standard and authoritative dataset for the zero pronoun resolution task.
For each of the anaphoric zero pronoun, which is predicted by the zero pronoun specific neural network (ZPSNN)~\cite{26}, the NRM generates a pronoun, of which the representation is an embedding vector.
The ZPSNN then extends its input with the recovered pronoun embedding and produces the zero pronoun resolution results.

To verify the effectiveness of the NRM for zero pronoun resolution, we compare the results of the ZPSNN and the ZPSNN+NRM with the results of the state-of-the-art (SOTA) approach~\cite{23} on zero pronoun resolution task.
The experimental results are shown in Table~\ref{tab6}.
As we use the same dataset for training, development and test with~\cite{23}, we directly show the original experimental results of~\cite{23} in Table~\ref{tab6}.

\begin{table}[htbp]
\caption{Experimental results on zero pronoun resolution. $\ast$ indicate that our approach is statistical significant over all the baselines (within $0.95$ confidence interval using the $t$-test).\label{tab6}}
\begin{center}{
\begin{tabular}{l|ccccccccc}
\hline
\multirow{3}{*}{\textbf{}} & \multicolumn{9}{c}{\textbf{Automatic Parsing \& Automatic AZP}}  \\
\cline{2-10}
& \multicolumn{3}{c}{\textbf{SOTA}} & \multicolumn{3}{c}{\textbf{ZPSNN}} & \multicolumn{3}{c}{\textbf{ZPSNN+NRM}} \\
\cline{2-10}
& \textbf{R} & \textbf{P} & \textbf{F} & \textbf{R} & \textbf{P} & \textbf{F} & \textbf{R} & \textbf{P} & \textbf{F} \\
\hline
overall & $19.6$ & $15.5$ & $17.3$ & $31.6$ & $24.5$ & $27.6$ & $\textbf{34.4}^{\ast}$ & $\textbf{24.8}^{\ast}$ & $\textbf{28.8}^{\ast}$  \\
\hline
NW & $11.9$ & $ 14.3$ & $13.0$ & $\textbf{20.0}$ & $\textbf{19.4}$ & $\textbf{19.7}$ & $\textbf{20.0}$ & $17.9$ & $18.9$  \\
MZ & $4.9$ & $4.7$ & $4.8$ & $12.5$ & $\textbf{11.9}$ & $12.2$ & $\textbf{14.3}$ & $11.5$ & $\textbf{12.8}$  \\
WB & $20.1$ & $14.3$ & $16.7$ & $34.4$ & $26.3$ & $29.8$ & $\textbf{38.6}$ & $\textbf{27.8}$ & $\textbf{32.3}$  \\
BN & $18.2$ & $\textbf{22.3}$ & $20.0$ & $24.6$ & $19.0$ & $21.4$ & $\textbf{26.4}$ & $20.3$ & $\textbf{23.0}$  \\
BC & $19.4$ & $14.6$ & $16.7$ & $35.4$ & $\textbf{27.6}$ & $31.0$ & $\textbf{41.6}$ & $27.4$ & $\textbf{33.1}$  \\
TC & $31.8$ & $17.0$ & $22.2$ & $50.7$ & $32.0$ & $39.2$ & $\textbf{52.1}$ & $\textbf{32.4}$ & $\textbf{39.9}$  \\
\hline
\end{tabular}}
\end{center}
\end{table}

The results on Table~\ref{tab6} show that by recovering the dropped pronouns for the anaphoric zero pronouns, the performance of the zero pronoun resolution can be further improved.
Meanwhile, the ZPSNN+NRM significantly outperforms the SOTA approach in statistics.
For further analysis, we will show the case study on the following section.

\subsection{Case Study on Zero Pronoun Resolution}

We compare the different cases that are produced by the ZPSNN and ZPSNN+NRM.
We find that the NRM can improve the performance of ZPSNN in the following kinds of cases.
\begin{itemize}
  \item The expected content for recovering a zero pronoun (antecedent) is an overt pronoun.

\begin{small}
\begin{tabular}{lr|llllll}
(9) & CN & 所以 & [$*AZP*$] & 有 & 一 & 个 & 想法 \\
& w2w & So & [$*AZP*$] & have & \multicolumn{2}{l}{a} & idea \\
\hline
& EN & \multicolumn{6}{p{6cm}}{So, [$we$] have an idea.} \\
\end{tabular}
\end{small}

Here, in the case (9), the $*AZP*$ is an anaphoric zero pronoun which should be referred to ``我们(We)''.
The NRM generates a dropped pronoun of ``我们(We)'' for the $*AZP*$.
Hence, the $*AZP*$ is correctly resolved.
While, without integrating the NRM, the $*AZP*$ is recovered as ``夜景(Night scene)'' by ZPSNN.

  \item The recovered dropped pronoun is a type indicator for the antecedent.

\begin{small}
\begin{tabular}{lr|llll}
(10) & CN & [$*AZP*$] & 以 & 信息化 & 带动 \\
& w2w & [$*AZP*$] & use & informationization & lead \\
& CN & 教育 & 的 & 现代化 & \\
& w2w & education & DE & modernization \\
\hline
& EN & \multicolumn{4}{p{7.3cm}}{[$The Ministry of Education$] simulates the modernization of education with informationization.} \\
\end{tabular}
\end{small}

Here, the NRM first generates a pronoun of ``它(It)'' for the $*AZP*$.
The system then chooses the candidate antecedents in the inanimate type.
Therefore, the $*AZP*$ is recovered as ``教育部(The Ministry of Education)'' by the ZPSNN+NRM system.
Whereas, the ZPSNN takes a name from the context as the antecedent of the $*AZP*$.

\end{itemize}

\section{Related Work}\label{rw}

The related studies with dropped pronoun recovery mainly focused on Empty Category (EC) recovery and zero anaphora resolution.
The EC recovery can be viewed as a previous step for DP recovery.
While zero anaphora resolution can be used either as features or an application for the DP recovery task.

\subsection{Empty Category Recovery}

Motivated by the government-binding theory, empty categories (ECs) are artificially annotated to explain specific language phenomenon in Penn Treebanks.

~\cite{4} proposed a log-linear based model to recover the ECs in Chinese Treebanks, by utilizing the lexical and syntactic features.
For training and testing, they encoded the surface node into two categories, EC and NEC (non-EC).
Hence, the recovery is actually a binary classification process.
However, the major drawback of this study is the overlook of structure and position information.
~\cite{11} presented a simple and highly effective EC recovery approach, which can fully integrate with state-of-the-art parser. The basic assumption is that the state-splitting of the parsing model will enable it to learn where to expect ECs to be inserted into the test sentences.
Experimental results indicated their superiority.
~\cite{3} described a novel approach to detecting ECs that represented in dependency parse trees.
They first converted the phrase structure into dependency structure.
The lexical and hierarchical features were then exploited for predicting both the position and type of ECs.
~\cite{5} explored a clause-level hybrid approach, which integrated the linear tagging and structure parsing information, to recovering ECs in Chinese.
They employed a higher level framework, semantic role labeling, to model the hybrid features.
Meanwhile, a comma disambiguation approach is also utilized to improve the performance of syntactic parsing and further affect the final results on EC recovery.
~\cite{2} proposed a structure learning based approach to recovering ECs for machine translation (MT) task.
3 major categories of features, lexical feature, syntactic feature and EC-specific feature, were taken into consideration.
They further validated that the proposed EC recovery approach can seamlessly integrate into the two state-of-the-art MT models and enhance the performance of MT systems.

Despite the similarity between the EC and DP recovery, the task of predicting and recovering DPs is more intricate due to its large modeling space.

\subsection{Zero Pronoun Resolution}

Zero anaphora is a pervasive phenomenon in Chinese, which is used to represent the co-reference between zero pronouns and their antecedents.
A zero pronoun (ZP) is a gap in a sentence which refers to an entity or event that can supply the necessary information for interpreting the gap.

~\cite{13} first performed the identification and resolution of Chinese anaphoric zero pronouns by using a machine learning approach.
They exploited two categories of features which can be summarized to intra- and inter-sentence features, including the position of the gap, syntactic roles, comma, clause, and the distances of zero pronouns and their antecedents in sentences.
Finally, a J48 decision tree model was employed to integrate these features for zero pronoun resolution.
~\cite{12} proposed a unified tree kernel based framework for Chinese zero pronoun resolution.
The main contribution of this study focused on measuring the similarity between zero pronouns and their antecedents on syntactic tree structure.
The task, then converted to classify the pair of zero pronoun and antecedent to anaphoric zero pronoun (AZP) and non-AZP by using the SVM classifier.
~\cite{14} further extended~\cite{13}'s work by involving more features and exploiting the co-reference links between zero pronoun and antecedent into a SVM$^{light}$ classifier.
Experimental results indicated that the extended approach outperformed the two state-of-the-art approaches~\cite{12,13} significantly.

A zero pronoun resolution task was also carried out on other pro-drop languages like Japanese and Korean.
~\cite{15} implemented the zero pronoun resolution task in a semantic role labeling framework.
They first transformed the dependency parsing trees of sentences into syntactic patterns.
A hybrid set of features, including lexical, grammatical, semantic and heuristic, as well as the syntactic patterns were uniquely integrated into the learning model.
~\cite{16} exploited the generated lexicalized case frames from large scale Web sentences as external knowledge.
They then proposed an example based log-linear model by integrating intra- and inter-sentence features for Japanese zero pronoun resolution.
~\cite{17} discussed the subject\/object drop phenomenon and the pattern found in the spoken and written text of child Korean.
Further, they analyzed the similar pro-drop phenomenon among seven languages, such as English, Italian, Portuguese, Chinese, Cantonese, and Japanese.
~\cite{18} presented a two step approach, which included the clause- and phrase-level antecedent detection, to resolving zero pronouns in Korean by reducing the number of candidate antecedents with syntactic structure information.
~\cite{22} proposed an unsupervised approach for Chinese zero pronouns resolution with language independent features.
They first predicted ten overt pronouns and then ranked the candidate antecedents.
The integer linear programming approach was then employed to enhance the performance of the ranking model.
They~\cite{23} further proposed an end-to-end unsupervised probabilistic model for Chinese zero pronoun resolution task.
They used a salience model to capture discourse information.
Experimental results showed that the proposed unsupervised model significantly outperformed the other approaches.

\section{Conclusion and Future Work}\label{con}

In this study, we present a novel neural network framework, namely NRM, for Chinese dropped pronoun recovery.
Experimental results show that the proposed NRM significantly outperforms the state-of-the-art approach in statistics.
Further, we integrate the recovered dropped pronoun into the zero pronoun resolution task.
Experimental results show that the performance of the zero pronoun resolution can be improved by recovering the dropped pronouns.

The future work will be carried out on the following:
First, we plan to explore more compatible neural networks to capture the context information and the compositional semantic of words.
Second, as one of the obstacles of the supervised learning approach is the lack of annotated training data, we plan to automatically construct or generate training data for dropped pronoun recovery.

\section*{Acknowledgement}

We thanks Linjie Wang for his help on data processing and running the SVM experiments.
This work is supported by the Fundamental Research Funds for the Central Universities (30620170037)

\section*{References}

\bibliography{elsarticle-template}
\end{CJK*}
\end{document}